\documentclass[preprint]{elsarticle}

\usepackage{hyperref}
\usepackage{multirow}
\usepackage{booktabs}
\usepackage{algorithm}
\usepackage{algorithmic}
\hypersetup{hypertex=true,linkcolor=green,anchorcolor=green,citecolor=green}

\begin{document}

\begin{frontmatter}

\title{UCP: Uniform Channel Pruning for Deep Convolutional Neural Networks Compression and Acceleration}

\author[mymainaddress]{Jingfei Chang}
\ead{cjfhfut@mail.hfut.edu.cn}

\author[mysecondaryaddress]{Yang Lu}

\author[mymainaddress]{Ping Xue}

\author[mysecondaryaddress]{Xing Wei}

\author[mysecondaryaddress]{Zhen Wei}


\address[mymainaddress]{School of Computer Science and Information Engineering, Hefei University of Technology, Hefei 230009, China}
\address[mysecondaryaddress]{Engineering Research Center of Safety Critical Industrial Measurement and Control Technology, Ministry of Education, Hefei University of Technology, Hefei 230009, China}

\begin{abstract}
Deep convolutional neural networks currently show the most advanced results in many artificial intelligence fields. However, with the continuous increase in the depth and width of CNNs, the number of parameters and floating-point operations (FLOPs) have also increased dramatically. To apply deep CNN to mobile terminals and portable devices, many scholars have recently worked on the compressing and accelerating deep CNN. Based on this, we propose a novel uniform channel pruning (UCP) method and the modified squeeze-and-excitation blocks (MSEB) is used to measure the importance of the channels in the convolutional layers. The unimportant channels, including convolutional kernels related to them, are pruned directly, which greatly reduces the storage cost and number of calculations. There are two types of residual blocks in ResNet. For ResNet with bottlenecks, we use the pruning method with traditional CNN to prune the 3$\times$3 convolutional layer in the middle of the blocks. For ResNet with basic blocks, we propose an approach to consistently prune all residual blocks in the same stage to ensure that the compact network structure is dimensionally correct. Considering that the network loses considerable information after pruning and that the larger the pruning amplitude is, the more information will be lost, we do not choose fine-tuning but retrain from scratch to restore the accuracy of the network after pruning. Finally, we verify our method on CIFAR-10, CIFAR-100 and ILSVRC-2012 for image classification. The results indicate that the performance of the compact network after retraining from scratch, when the pruning rate is small, is better than the original network. Even when the pruning amplitude is large, the accuracy can be maintained or decreased slightly. On the CIFAR-100, when reducing the parameters and FLOPs up to 82\% and 62\% respectively, the accuracy of VGG-19 even improved by 0.54\% after retraining.
\end{abstract}


\end{frontmatter}

\section{Introduction}
In recent years, with the rise of deep learning \citep{ISI:000237698100002,ISI:000239308600057,ISI:000355286600030}, deep neural networks, which have produced more advanced results than other methods, have been widely used in various fields. Convolutional neural networks \citep{ISI:000287216000064}, because of their translation invariance and parameter sharing properties, have made great breakthroughs in the fields of speech, image and video. However, with the gradual advancement of research, the depth and width of convolutional neural networks have been increasing, from traditional CNNs such as LeNet \citep{INSPEC:6081653}, AlexNet \citep{INSPEC:17133663}, ZFNet \citep{DBLP:conf/eccv/ZeilerF14} and VGGNet \citep{2014arXiv1409.1556S} to GoogLeNet \citep{DBLP:conf/cvpr/SzegedyLJSRAEVR15}, ResNet \citep{DBLP:conf/cvpr/HeZRS16} and DenseNet \citep{DBLP:conf/cvpr/HuangLMW17} with special structures. The depth of the network currently ranges from the first few layers to more than 200 layers. As CNNs become surprisingly complex, their representation capabilities continue to improve. Nevertheless, the rapid development of convolutional neural networks has greatly benefited from the advent of GPUs, TPUs, cloud processors, and extremely large datasets, which makes it difficult for some CNNs that perform well yet have very deep structures to be deployed on mobile devices with limited memory, computing capabilities and high real-time requirements.

To solve this problem, many scholars have devoted themselves to compressing and accelerating convolutional neural networks. Many methods aiming to reduce the number of network parameters and floating-point operations (FLOPs) have been proposed based on the parameter redundancy and structural characteristics of CNNs and have made significant progress. These methods are mainly divided into two categories. First, some hardware implementation methods \cite{ISI:000402739900008,GONG2017384} are proposed for the matrix operations, and dedicated hardware accelerators \cite{ISI:000360535000019,ISI:000485228900009} are designed to accelerate the training and inference. Second, some algorithms, such as network pruning, quantization, low-rank decomposition, and knowledge distillation, have been proposed to reduce the network's memory footprint and FLOPs. Parameter quantization is divided into ordinary quantization\cite{2016arXiv160606160Z} and extreme binary neural networks\cite{ISI:000450913100015,ISI:000389385100032,INSPEC:18187535,LI2020}. Parameter quantization can speed up the calculations but cannot reduce the number of parameters, and when the network is more complex, the inference accuracy is lower. The low-rank factorization operation\cite{ISI:000452647102105,ISI:000425646500006} can achieve network sparseness and directly compress and accelerate the network, but additional calculations are introduced in the implementation process, which is not conducive to reducing the FLOPs. Knowledge distillation\cite{INSPEC:16231013,INSPEC:16791229} can make deep networks narrower, but the similarity required for the two network tasks is higher, and actual performance may be worse. Some people have combined multiple methods to compress and accelerate convolutional neural networks. \cite{ISI:000498677600008} combined low-rank decomposition (LRD) and knowledge transfer (KT) to accelerate and compress CNN. \cite{DING2019106957} used teacher-student learning and Tucker decomposition methods to reduce model size and runtime latency. \cite{XU2019272} proposed a general framework of architecture distillation, namely LightweightNet, to accelerate and compress convolutional neural networks.

In this paper, we propose a novel uniform channel pruning (UCP) method that integrates the modified squeeze-and-excitation blocks (MSEB). We use the outputs of the middle activation layer in MSEB as an evaluation index of the output channel importance in the convolutional layer. A coefficient is set as a hyperparameter to determine which channels need to be pruned, and all related filters are cropped. We prune the traditional convolutional neural network VGGNet and ResNet with special structures. VGGNet can directly prune the channels of each layer based on our method, but ResNet cannot for a particular residual block. The ResNet with the basic residual block, due to its structure, limits the dimensions of the input and output and cannot individually prune each residual block. We use the importance evaluation index to uniformly cut the layers with the same dimensions in a certain stage to ensure the correctness of its structure. For ResNet with bottlenecks, we only cut the 3$\times$3 convolutional layers in the middle of the residual blocks. We found that when the pruning amplitude is large, due to the loss of parameter information, it cannot be restored to competitive accuracy through fine-tuning. Therefore, we retrain the compact network from scratch to restore the performance after pruning. Extensive experiments show that our method can greatly reduce the parameters and FLOPs of deep convolutional neural networks, and the compact network has equivalent or even better accuracy than the original network.

The main contributions of our work are as follows:
\begin{itemize}
\item We propose a new MSEB-based convolutional neural network pruning method. The sigmoid activation in the middle of the MSEB module is used to evaluate the importance of the intermediate feature maps generated by the CNNs for image classification tasks. Unimportant feature maps are clipped with their associated filters to compress and accelerate the network.

\item We design a method for setting the network pruning threshold based on the characteristics of the MSEB module output in the convolutional neural networks to compress the CNNs with different proportions. For ResNets, we propose a consistent compression strategy in the same stage.

\item We perform extensive experiments based on our method on three general image classification datasets CIFAR-10, CIFAR-100 and ILSVRC-2012, and compare our method with some existing advanced methods. 
\end{itemize}

\section{Ralated Work}
Since the emergence of convolutional neural networks in image classification and other fields, many scholars have begun to study network compression and acceleration and have proposed many excellent methods. Convolutional neural network pruning has received extensive attention and research as one of the most popular methods. The network pruning is divided into filter-level and channel-level.

\paragraph{\textbf{Filter Level}} \cite{ISI:000450913101044} presented the three-step pipeline method of training, pruning, and fine-tuning to achieve weight-level trimming of convolutional neural networks. \cite{2016arXiv160808710L} used the L1 norm of the filters as an evaluation index of the importance of weights to prune filters from CNNs that are identified as having a small effect on the output accuracy. \cite{INSPEC:16308044,2017arXiv170605791L} used the sparseness and the entropy in the feature maps to perform network pruning. \cite{DBLP:conf/cvpr/YangCS17} proposed a trimming method based on energy efficiency.  SFP (Soft Filter Pruning) \cite{DBLP:conf/ijcai/HeKDFY18} used the L2 norm for pruning, and allow the pruned filters to be updated during the next training procedure. These training process is continued until converged. \cite{ISI:000489763000018} trimmed the filters based on the statistical data calculated by the next layer instead of the current layer, and a more accurate group convolution scheme "gcos" (Group COnvolution with Shuffling) to further reduced the size of the pruned model. \cite{INSPEC:19070985} developed an algorithm named as COP, which can detect the redundant filters efficiently and enables the users to customize the compression according to their preference (smaller or faster). \cite{ASHOURI201956} proposed and evaluated three model-independent methods for sparsification of model weights and explore retraining-free pruning of CNNs.

\paragraph{\textbf{Channel Level}} \cite{DBLP:conf/iccv/LiuLSHYZ17} leveraged the scaling layers in batch normalization to effectively identify and prune unimportant channels in the network. \cite{DBLP:conf/iccv/HeZS17} proposed an iterative two-step algorithm, for the pretrained CNN models, to effectively prune each layer using LASSO regression and least square reconstruction algorithm, and then generalized it to multiple layers and multi-branch situation. Different from previous methods. \cite{INSPEC:19263174} proposed Centripetal SGD, which can train several filters to collapse into a single point in the parameter hyperspace to trim the network with no performance loss. \cite{TIAN201845} proposed a structured filter-level pruning approach based on Fisher LDA. \cite{DBLP:conf/cvpr/LinJYZCYHD19} proposed structured sparsity regularization (SSR), to simultaneously speedup the computation and reduce the memory overhead of CNNs. \cite{2019arXiv190908174Y} proposed a global filter pruning algorithm, which transforms a vanilla CNN module by multiplying its output by the channel-wise scaling factors. \cite{SINGH2020103857} added a fatuous auxiliary task for filter pruning. \cite{XIE2020} analyzed most of existing pruning methods mainly focused on classification and few of them conducted systematic research on object detection. Therefore, based on discrimination-aware channel pruning (DCP) which is state-of-the-art pruning method for classification, they proposed a localization-aware auxiliary network to find out the channels with key information for classification and regression so that they can conduct channel pruning directly for object detection, which saved lots of time and computing resources. \cite{LUO2020107461} proposed an end-to-end trainable system by combine channel pruning and model fine-tuning, and designed an effective channel selection layer AutoPruner to automatically find less important channels and convolution kernels for pruning. \cite{INSPEC:18173228} proposed AutoML for Model Compression(AMC), which used reinforcement learning to provide a model compression strategy.

Our method can compress and accelerate the CNN at the channel level based on the redundancy of the parameters and FLOPs without any sparseness introduced. Our method can uniformly prune the network without layer-wise training which will increase training costs additionally. Finally, our method does not require the assistance of additional sparse matrix operations and acceleration libraries.

\section{Network Pruning}

\subsection{Modified Squeeze and Excitation Blocks}

\begin{figure}[h]
	
	\centering
	
	\includegraphics[height=2.4cm,width=11.7cm]{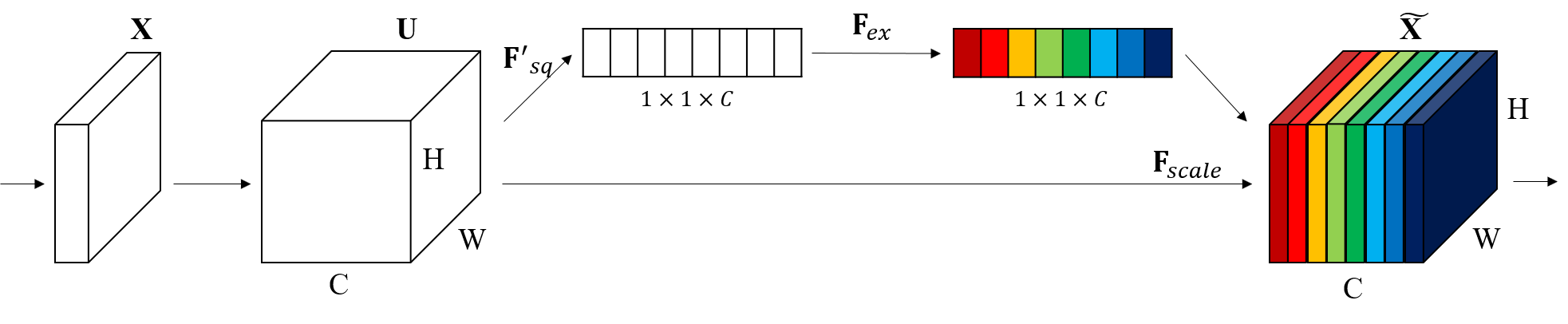} 
	
	\caption{Squeeze-and-Excitation module}
	\label{img1}
	
\end{figure}

The SENet presented by \cite{DBLP:conf/cvpr/HuSS18} with the SE blocks, which can be directly integrated into various convolutional neural networks, ranked first in the ILSVRC 2017 classification competition. The original SE module is shown in Figure \ref{img1}. For any given input X, it is transformed into a three-dimensional feature map ${\rm{U}} \in {R^{H \times W \times C}}$ through the convolution operation. First, the feature map is compressed, and the H $\times$ W dimensional data of each channel is aggregated into a tensor of 1 $\times$ 1 $\times$ C through global average pooling and then transforms it into a tensor of 1 $\times$ 1 $\times$ C/r through a fully connected layer followed by ReLU. After that, a fully connected layer was used to produce a tensor of  1 $\times$ 1 $\times$ C, and then the C results are restricted between 0 and 1 based on the sigmoid. Finally, the original feature map is multiplied by this tensor as the input of the next layer.

We use the tensor of 1 $\times$ 1 $\times$ C in the SE module after Sigmoid as the evaluation index of the importance for each channel in the feature map. Considering the feature map have both positive and negative values, if the original global average pooling is used, the positive and negative terms will counteract each other. In order to retain as much information as possible in the feature map, we take the absolute value of each item of the feature map in the original SE module and then followed by the global average pooling. The fully connected layers in the SE module is replaced with convolutional layers, which is more conducive to capture the spatial information in the feature map, and our MSEB module is obtained. In our method, given the feature map ${\rm{U = }}\left[ {{{\rm{u}}_1},{{\rm{u}}_2}, \ldots,{{\rm{u}}_C}} \right]$, where ${{\rm{u}}_c} \in {R^{H \times W}}$, the c-th element in the compressed tensor ${\rm{z}} \in {R^C}$ of 1 $\times$ 1 $\times$ C is calculated as follows:

\begin{equation}
{{z_c} = {{{\bf{F}}'_{sq}}}\left( {{{\mathop{\rm u}\nolimits} _c}} \right) = \frac{1}{{H \times W}}\sum\limits_{i = 1}^H {\sum\limits_{j = 1}^W {\left| {{u_c}\left( {i,j} \right)} \right|} }}.
\end{equation}
The operation of the excitation layer after compressed is as follows:

\begin{equation}
{{\rm{s}} = {{\bf{F}}_{ex}}\left( {{\mathop{\rm z}\nolimits} ,{\mathop{\rm W}\nolimits} } \right) = \sigma \left( {{g}\left( {{\mathop{\rm z}\nolimits} ,{\mathop{\rm W}\nolimits} } \right)} \right) = \sigma \left( {{{\mathop{\rm W}\nolimits} _2}\delta \left( {{{\mathop{\rm W}\nolimits} _1}{\mathop{\rm z}\nolimits} } \right)} \right)},
\end{equation} 
where, $\sigma$ refers to sigmoid activation, $\delta$ refers to ReLU and weights ${{\mathop{\rm W}\nolimits} _1} \in {R^{\frac{C}{{\rm{r}}} \times C}}$, ${{\mathop{\rm W}\nolimits} _2} \in {R^{C \times \frac{C}{{\rm{r}}}}}$. Finally, we use the above activation s and input feature map U to obtain the final output:

\begin{equation}
{{\widetilde {\rm{x}}_c} = {{\bf{F}}_{scale}}\left( {{\mathop{\rm u}\nolimits} {}_c,{s_c}} \right) = {s_c}{\mathop{\rm u}\nolimits} {}_c},
\end{equation}
where, $\widetilde {\mathop{\rm X}\nolimits}  = \left[ {{{\widetilde {\mathop{\rm x}\nolimits} }_1},{{\widetilde {\mathop{\rm x}\nolimits} }_2}, \ldots ,{{\widetilde {\mathop{\rm x}\nolimits} }_C}} \right]$ and ${{\bf{F}}_{scale}}\left( {{\mathop{\rm u}\nolimits} {}_c,{s_c}} \right)$ refers to Channel-wise product of ${s_c}$ and input feature map ${{\rm{u}}_c} \in {R^{H \times W}}$.

\subsection{Pruning Strategy on Channels}

We design a network pruning method for traditional convolutional neural networks and ResNets with special residual structures. The output tensor of 1 $\times$ 1 $\times$ C of the MSEB module is used as the importance evaluation index of the feature maps. The unimportant feature maps and all the filters related to them are pruned together to adequately reduce the parameters and FLOPs. The MSEB module obtain C values between 0 and 1 for the C-dimensional feature maps of each convolutional layer. According to the characteristics of the output results of the MSEB module in the experiment, we propose a pruning threshold setting method as follows:

\begin{equation}
{Thre =(1 \pm \lambda) \frac{1}{C} \sum_{i=1}^{C} {f}^{(i)}_{se} (\cdot)},
\end{equation}
where, $\lambda = 1\times10^{-\beta}$, $i$ is the index of the feature map, and $Thre$ refers to the pruning threshold. The feature maps corresponding to the output values that are smaller than the threshold will be deleted. As shown in Figure \ref{img2}, The left is the original CNN, and the middle is a pruning schematic diagram of the 8-channel convolutional layer. According to our method, the original 8-channel output feature maps are trimmed according to the threshold. The channels pruned and relevant filters are displayed in the upper right corner. In the experiment, we determine the number of clipped channels by controlling $\beta$ and the sign of $\lambda$. In the rest of this paper, unless specified, the sign of $\lambda$ is negative by default.

\begin{figure}[h]\label{img2}
	
	\centering
	
	\includegraphics[height=5.3cm,width=12cm]{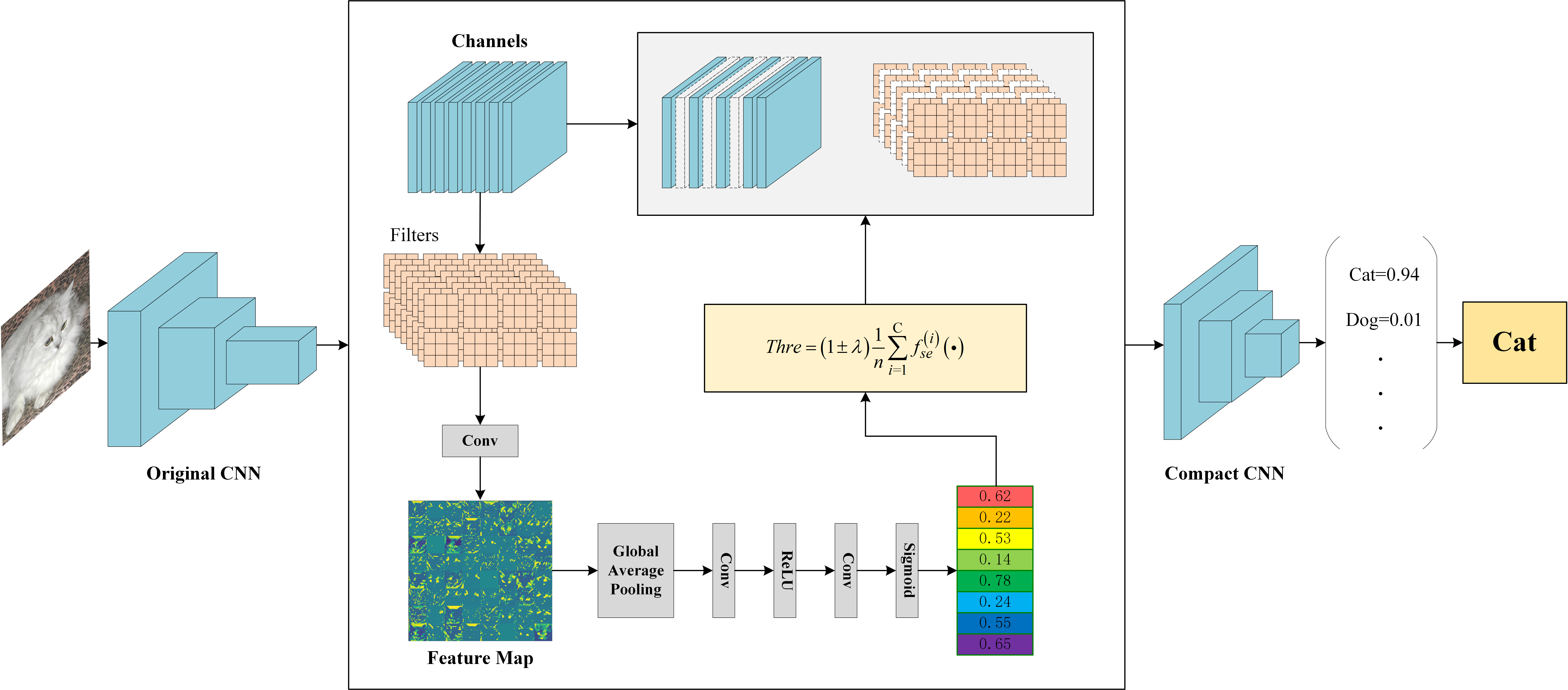} 
	
	\caption{Feature map pruning schematic diagram of traditional convolutional neural networks based on our UCP}
	\label{img2}
	
\end{figure}

After the feature maps are pruned, the filters connected to them are also deleted so that the network is further compressed. Unlike some existing layer-wise network pruning methods, we firstly use the pretrained CNN to generate the MSEB output for each layer and then set a huperparameter $\beta$. Finally, we clip all channels in each layer that are smaller than the threshold to obtain the final compact network. Our method prunes all layers at the same time. According to the specific characteristics of the feature maps, the number of clipping channels of each layer is different, even under the same value of $\beta$. In this way, each layer can be controlled by $\beta$ to prune as many of the less important channels as possible. In the experiment, some output of the MSEB layer are 0.5 due to the sigmoid. When the pruning amplitude is large, we directly cut these layers in half.

The main difference between ResNet and the traditional network is its special residual module, which has two types. One is composed of basic residual blocks that contain two 3 $\times$ 3 convolutional layers. The other is composed of bottleneck residual blocks, which include two 1 $\times$ 1 convolutional layers and a 3 $\times$ 3 convolutional layer between them.

\begin{figure}[h]
	
	\centering
	
	\includegraphics[height=6cm,width=12cm]{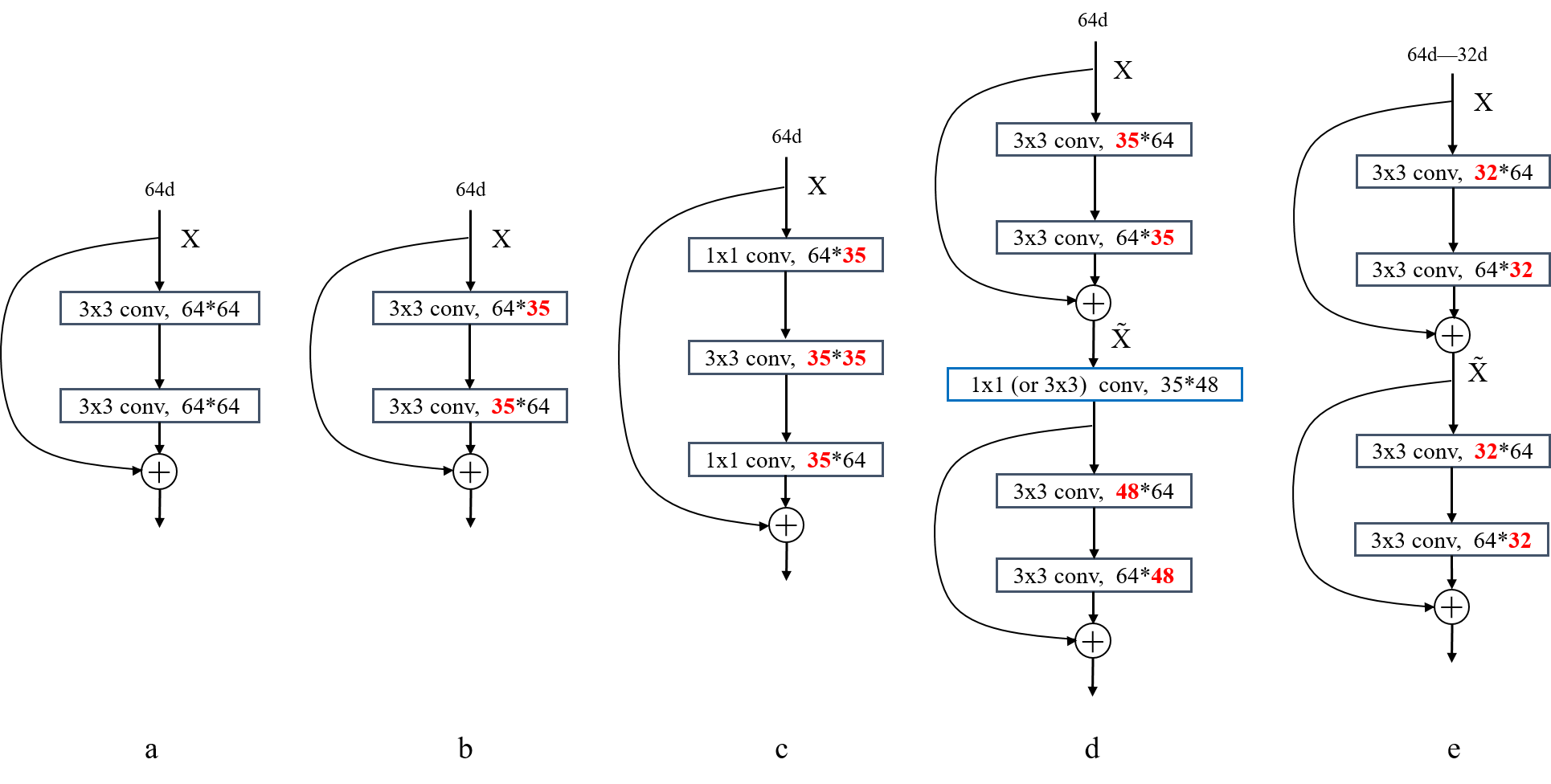} 
	
	\caption{Pruning methods for ResNet. $a$ is an original 64-dimensional residual block; $b$ shows the existing methods to prune the intermediate channel (feature map) for ResNet with the basic residual block; $c$ is the method for pruning the ResNet with the bottleneck residual block. We only prune the intermediate 3 $\times$ 3 convolutional layer; $d$ is a method of adding a 1 $\times$ 1 convolutional layer between the residual blocks to clip the input and output channels of each residual block. We abandoned this method after the experiment; $e$ is the pruning method we finally adopted for ResNet with basic residual blocks. We adopt the same clipping range; in this figure, the 64-dimensional residual blocks are pruned to 32 dimensions in one stage, and prune the input and output channels of all residual blocks are pruned to the same dimension.}
	\label{img3}
	
\end{figure}

For the first type of ResNet, due to its limitation of special residual structure, only the intermediate output channels of the residual blocks can be pruned. All the existing cutting methods use different pruning strategies to perform the intermediate channels to protect the structure. We attempt to prune the input channels and output channels of the residual block separately so that the flexibility of pruning is greater. We add a 1 $\times$ 1 convolutional layer between every two residual blocks, as shown in Figure \ref{img3}(d), to ensure that the input and output channels of the residual blocks can be pruned separately without affecting the normal dimensional restriction of the ResNet and without introducing too many extra parameters and FLOPs for the network. We reconstruct ResNet-56 according to the above method and perform experiments on CIFAR-10. Under the same experimental environment and parameter configuration, the classification accuracy of the network constructed in this way decreases by approximately 1.7\% compared to the original ResNet. After adding the MSEB module to the reconstructed ResNet-56, the accuracy decreased even more seriously, by approximately 2.5\%. It causes the performance deteriorating even more severely after pruning. Therefore, we propose another pruning method for ResNet. Since each stage of the ResNet contains a certain number of residual blocks, the input and output dimensions of these residual blocks are consistent. We take the same trimming range for all residual blocks in this stage, and only prune the input and output channels and the corresponding filters, reserving the intermediate feature maps.

In the implementation process, we add the MSEB module after the last convolutional layer of each residual block, using the same method to train ResNet. Then, we collect and analyze the output values of the sigmoid in the MSEB module. We determine the consistent trimming range at each stage through its overall characteristic and finally experimentally verify that the pruning strategy is effective and better than some existing network pruning methods.

For the second type of ResNet that consisting of bottleneck residual blocks, since the 1 $\times$ 1 convolutional layers have considerably fewer parameters and FLOPs than the 3 $\times$ 3 layers, we perform channel-wise pruning on the middle 3 $\times$ 3 convolutional layers and delete the filters related to them. In the implementation, we embed the MSEB module behind the 3 $\times$ 3 convolutional layer in the middle of the bottleneck residual block. After training the ResNet to convergence, we set a hyperparameter to prune the network according to the traditional CNN pruning method introduced in the previous section.

\subsection{Implementation details}

\begin{figure}[h]
	
	\centering
	
	\includegraphics[height=8cm,width=12cm]{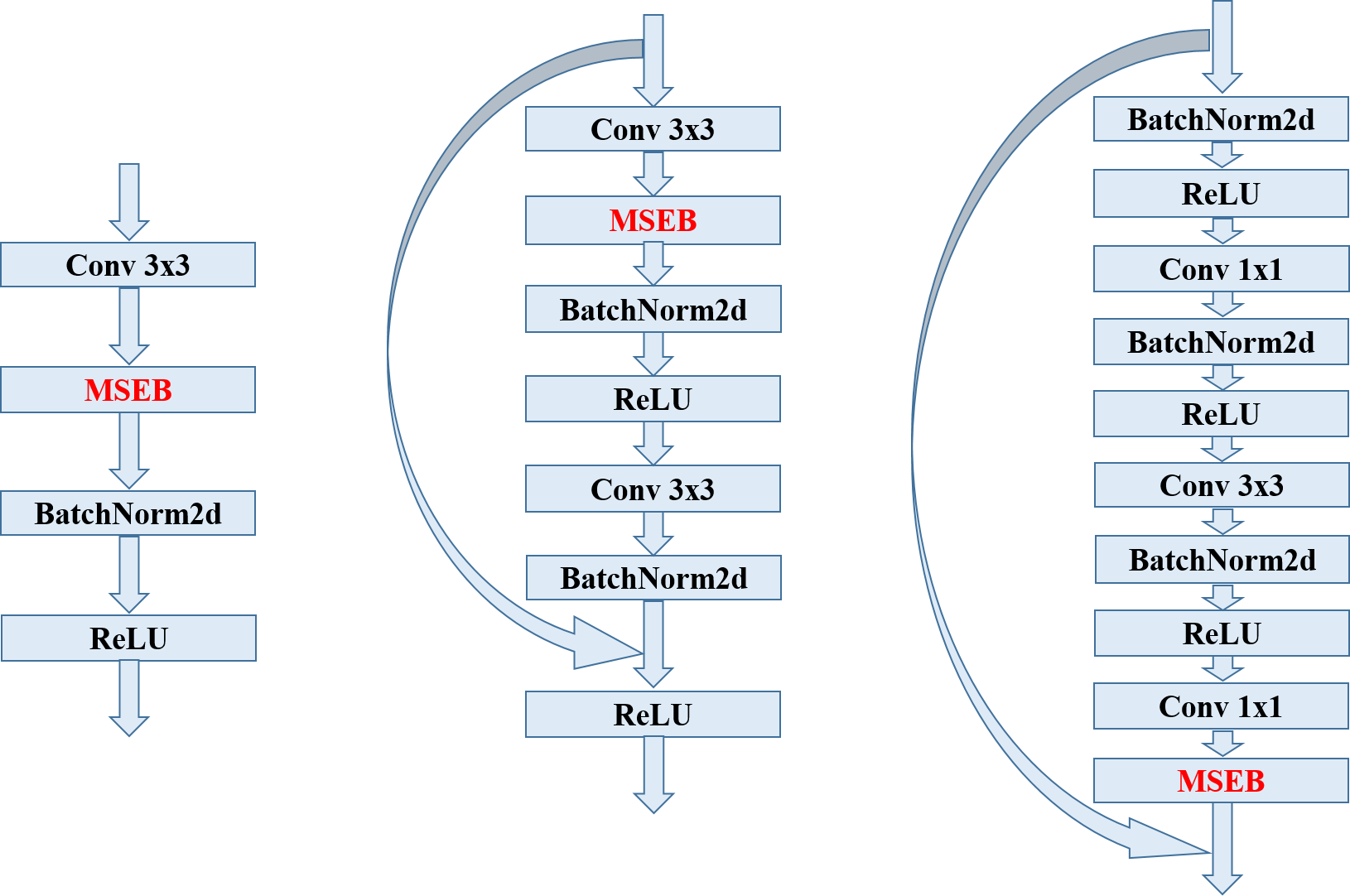} 
	
	\caption{The location of the MSEB module embedded in VGGNet, ResNet and PreResNet. The left picture shows a part of the VGGNet added the MSEB module. The middle picture shows the MSEB module added to the basic residual block, and the right picture shows the bottleneck with MSEB module.}
	\label{img4}
	
\end{figure}

We apply our method to classical network compression. Given a convolutional neural network, we embed the MSEB module into a specified position. For VGGNet, an MSEB module is added directly after each convolutional layer. In order to maximize the role of the MSEB module, we place it before the ReLU activation layer. For the ResNet, different embedding methods of the MSEB module are performed according to the type of the residual block. For the basic block, each block contains two 3 $\times$ 3 convolutional layers. We want to prune the middle channel, so the MSEB module is placed after the first convolution layer and before the Batch Normalization and the ReLU. But for the bottleneck, each block contains three convolutional layers. In order to better measure the importance of each channel of the entire residual block, we add the MSEB module after the third convolutional layer. The detailed network structure is shown in Figure \ref{img4}.

Given a dataset, we first train the convolutional neural network with the MSEB module. After converged, the model with the highest classification accuracy is selected as our target model, and we extract the output value of the MSEB module. We determine the hyperparameter according to the output value characteristics of the middle layer in the MSEB module, and then calculate the pruning threshold. In the training process, we use the following cross entropy as the loss function:

\begin{equation}
{L =  - \sum\limits_{i = 1}^N {\left[ {{y^{\left( i \right)}}log{{\widehat y}^{\left( i \right)}} + \left( {1 - {y^{\left( i \right)}}} \right)log\left( {1 - {{\widehat y}^{\left( i \right)}}} \right)} \right]}},
\end{equation}
where  ${y^{\left( i \right)}}$ and ${\widehat y^{\left( i \right)}}$ refer to the real probability and estimated probability of the \textit{i}-th category respectively. \textit{N} is the total number of image categories. In order to control the magnitude of the weights in the back-propagation and avoid gradient explosion, we introduce $penalty$ in the loss function, which is also conducive to increasing the sparsity of the weights and improving the generalization ability. For the penalty term, we choose L2 norm. The final loss function is as follows:

\begin{equation}
{L{\rm{oss}} = L{\rm{ + }}penalty{\rm{ = }} - \sum\limits_{i = 1}^N {\left[ {{y^{\left( i \right)}}log{{\widehat y}^{\left( i \right)}} + \left( {1 - {y^{\left( i \right)}}} \right)log\left( {1 - {{\widehat y}^{\left( i \right)}}} \right)} \right]} {\rm{ + }}\frac{\lambda }{{2n}}\sum\limits_{i = 1}^n {{w_i}^2}}.
\end{equation}
In the back-propagation, we choose the stochastic gradient descent with momentum. Selecting a small batch of \textit{m} samples $\left\{ {{x^{\left( 1 \right)}},{x^{\left( 2 \right)}}, \ldots ,{x^{\left( m \right)}}} \right\}$ from the training set, the corresponding target real probability is ${y^{\left( i \right)}}$. We calculate the gradient estimate of the parameter after the forward propagation:

\begin{equation}
{g = \frac{1}{m}{\nabla _W}\sum\limits_{i = 1}^m {Loss\left( {f\left( {{x^{\left( i \right)}};W} \right),{y^{\left( i \right)}}} \right)}},
\end{equation}
where, ${f\left( {{x^{\left( i \right)}};W} \right)={{\widehat y}^{\left( i \right)}}}$. Then the gradient update method is as follows:

\begin{equation}
{\begin{array}{l}
	{\omega _{t + 1}} = {\omega _t} + {v_t} = {\omega _t} + \alpha {v_{t - 1}} - \eta {g_t}\\
	= {\omega _t} + \alpha {v_{t - 1}} - \eta \frac{1}{m}{\nabla _W}\sum\limits_{i = 1}^m {Loss\left( {f\left( {{x^{\left( i \right)}};W} \right),{y^{\left( i \right)}}} \right)} 
	\end{array}
},
\end{equation}
where, $v$ is the initial gradient accumulation, ${v_0} = 0$, $\alpha$ is the coefficient of momentum, and $\eta $ is the learning rate. We use this method to train the CNNs on the image classification dataset. Our pruning algorithm is show as Algorithm \ref{alg1}.

\begin{algorithm}[htbp]
	\caption{UCP: uniform channel pruning algorithm}
	\label{alg:algorithm}
	\textbf{Input}: Dataset ${\bf{D}}{\rm{ = }}\left\{ {\left( {{x^0}{\rm{,}}{{\rm{y}}^0}} \right),\left( {{x^1}{\rm{,}}{{\rm{y}}^1}} \right), \ldots ,\left( {{x^M}{\rm{,}}{{\rm{y}}^M}} \right)} \right\}$, feature map  ${\rm{U = }}\left[ {{{\rm{u}}_1},{{\rm{u}}_2}, \ldots ,{{\rm{u}}_C}} \right]$\\
	\textbf{Parameter}: $\lambda$, $\alpha$, $\eta$, $\beta$\\
	\textbf{Output}: Impact network
	\begin{algorithmic}[1] 
		\FOR {$epoch=1$ to $K$}
		\STATE forward propagation: $L{\rm{oss}} = L{\rm{ + }}\frac{\lambda }{{2n}}\sum\limits_{i = 1}^n {{w_i}^2}$.
		\STATE back propagation: ${\omega _{t + 1}} = {\omega _t} + \alpha {v_{t - 1}} - \eta \frac{1}{m}{\nabla _W}\sum\limits_{i = 1}^m {Loss\left( {f\left( {{x^{\left( i \right)}};W} \right),{y^{\left( i \right)}}} \right)}$.
		\ENDFOR
		\STATE $\lambda  = 1 \times {10^{ - \beta }}$
		\STATE ${Thre =(1 \pm \lambda) \frac{1}{C} \sum_{i=1}^{C} {f}^{(i)}_{se} (\cdot)}$	
		\FOR {$i=1$ to $C$}
		\IF {${s_i} < Thre$}
		\STATE Pruning the i-th channel and filters related
		\ENDIF
		\ENDFOR
		\STATE \textbf{return} solution
	\end{algorithmic}
    \label{alg1}
\end{algorithm}

After the network is pruned, most of the existing methods inherit the weights and biases from the original pretrained network and restore the performance of the compact network as much as possible by fine-tuning \citep{2016arXiv160808710L,DBLP:conf/ijcai/HeKDFY18,DBLP:conf/iccv/LiuLSHYZ17,DBLP:conf/iccv/HeZS17,ISI:000489763000018}. However, when the network pruning amplitude is large, considerable parameter information is lost, and the performance recovery is not obvious after fine-tuning, so the actual performance of the compact network cannot be fully displayed. \cite{liu2018rethinking} exposed a surprising character in structured network pruning: the performance of the compact network by fine-tuning with inherited parameters is not better than training from scratch.

The experimental results obtained by using VGG-16 on the CIFAR-10 further verify the correctness of the conclusions in \cite{liu2018rethinking}. To adequately reflect the performance of the compact network, we retrain the pruned networks from scratch and keep the FLOPs before and after pruning consistent in the experiments. Specifically, we multiply the training epochs of the original network by the FLOP compression rate as the retraining epochs of the pruned network.

\section{Experiments and Results}

To prove that the proposed method is effective, we prune VGGNet, ResNet, and PreResNet, followed by performing experiments on the CIFAR-10, CIFAR-100 and ILSVRC-2012 for image classification. We implement our experiments on PyTorch. The network pretraining and hyperparameter settings use the method presented in [10] to obtain better benchmark accuracy and can be compared fairly with existing network pruning methods. The learning rate is initially set to 0.1 and then decreased by a factor of 10 on half and three-quarter epochs. We train 160 epochs on the CIFAR-10 and CIFAR-100 and 100 epochs on the ILSVRC-2012. We use SGD with a mini-batch size of 64. We use a weight decay of 0.0001 and a momentum of 0.9. The reduction in the MSEB module is set to 16, and the BN layer is added between the convolutional layer and the activation layer. In order to exclude the influence of the experimental equipment and frame on the final classification accuracy, we use the percentage of accuracy reduction of the compact network relative to the baseline, and for the completeness of the experimental data, we also show all values of parameters and FLOPs.

\begin{figure}[h]
	
	\centering
	
	\includegraphics[height=7cm,width=14cm]{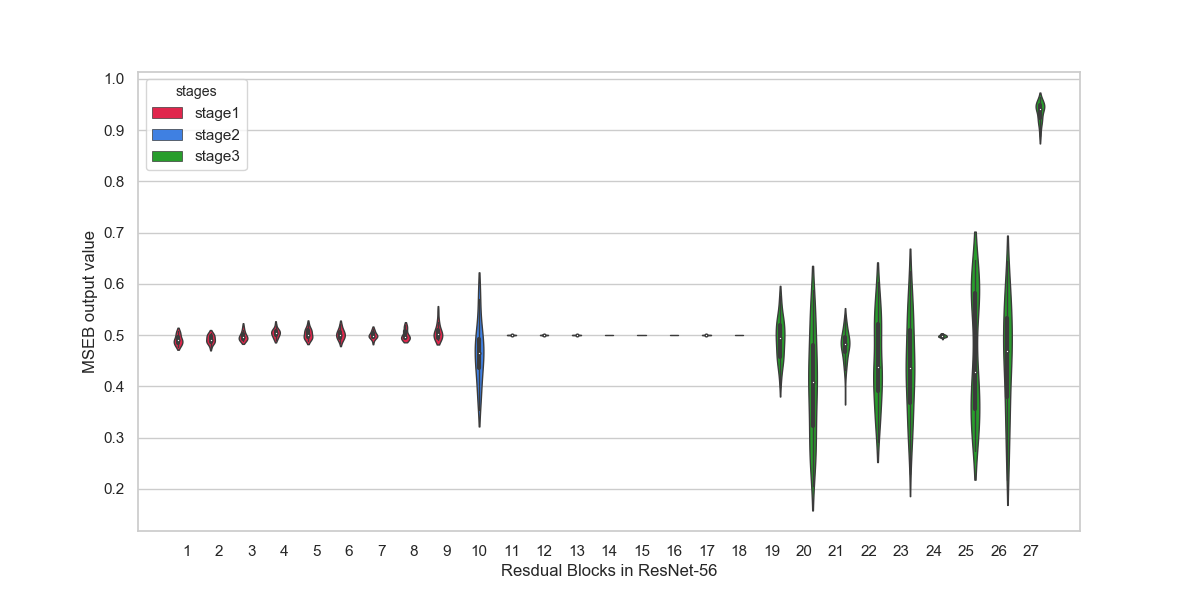} 
	
	\caption{Statistical characteristics of the MSEB output in each residual block of the ResNet-56 network on the CIFAR-10}
	\label{img5}
	
\end{figure}

\begin{table*}[htbp]
	
	\centering
	\renewcommand{\arraystretch}{2}
	\resizebox{\textwidth}{!}{
		\begin{tabular}{|c|c|c|c|c|c|c|c|c|}
			
			\hline 
			Model             & S/F & Baseline & Acc(\%)  & Acc.Drop(\%) & Parameters & Pruned(\%) & FLOPs & Pruned(\%) \\ 
			\hline 
			VGG-16              & $-$ & 93.55 & $-$     & $-$     & $1.50 \times 10^{7}$ & $-$     & $3.14 \times 10^{8}$ & $-$ \\ 
			
			MSEBVGG-16            & $-$ & 93.84 & $-$     & $-$     & $1.52 \times 10^{7}$ & $-$     & $3.15 \times 10^{8}$ & $-$ \\ 
			
			\multirow{2}*{VGG16-prund(4)}  & F & 93.55 & 92.18 & 1.37  & $8.73 \times 10^{6}$ & 41.8  & $1.58 \times 10^{8}$ & 49.7 \\
			
			& S & 93.55 & 93.65 & -1.00  & $8.73 \times 10^{6}$ & 41.8  & $1.58 \times 10^{8}$ & 49.7 \\
			
			\multirow{2}*{VGG16-prund(8)}  & F & 93.55 & 91.57 & 1.98  & $4.79 \times 10^{6}$ & 68.1  & $9.96 \times 10^{7}$ & 68.3 \\ 
			
			& S & 93.55 & \textbf{93.75} & \textbf{-0.20} & $\textbf{4.79} \times \textbf{10}^{\textbf{6}}$ & \textbf{68.1}  & $\textbf{9.96} \times \textbf{10}^{\textbf{7}}$ & \textbf{68.3} \\ 
			
			\multirow{2}*{MSEBVGG16-prund(4)}    & F & 93.55 & 92.35 & 1.20 & $8.86 \times 10^{6}$ & 40.9  & $1.59 \times 10^{8}$ & 49.4 \\ 
			
			& S & 93.55 & 93.71 & -0.16 & $8.86 \times 10^{6}$ & 40.9  & $1.59 \times 10^{8}$ & 49.4 \\ 
			
			\multirow{2}*{MSEBVGG16-prund(8)}    & F & 93.55 & 91.53 &  2.02 & $4.86 \times 10^{6}$ & 67.6  & $9.99 \times 10^{7}$ & 68.2 \\ 
			
			& S & 93.55 & 93.50 &  0.05 & $4.86 \times 10^{6}$ & 67.6  & $9.99 \times 10^{7}$ & 68.2 \\ 
			
			\cite{2016arXiv160808710L}    & F & 93.25 & 93.40 & -0.15 & $5.40 \times 10^{6}$ & 64.0  & $2.06 \times 10^{8}$ & 34.2 \\ 
			
			\cite{liu2018rethinking}        & S & 93.63 & 93.78 & -0.15 & $5.40 \times 10^{6}$ & 64.0  & $2.06 \times 10^{8}$ & 34.2 \\ 		
			
			\hline 
			\hline
			ResNet-56             & $-$ & 93.15 & $-$     & $-$     & $8.53 \times 10^{5}$ & $-$     & $1.27 \times 10^{8}$ & $-$ \\ 
			
			MSEBResNet-56           & $-$ & 93.44 & $-$     & $-$     & $8.59 \times 10^{5}$ & $-$     & $1.27 \times 10^{8}$ & $-$ \\ 
			
			ResNet-56(8-32-32)    & S & 93.15 & \textbf{93.28} & \textbf{-0.13} & $\textbf{5.15} \times \textbf{10}^{\textbf{5}}$ & \textbf{39.6}  & $\textbf{8.47} \times \textbf{10}^{\textbf{7}}$ & \textbf{33.3} \\ 
			
			\cite{2016arXiv160808710L}        & F & 93.04 & 93.06 & -0.02 & $7.30 \times 10^{5}$ & 13.7  & $9.09 \times 10^{7}$ & 27.6 \\ 
			
			\cite{liu2018rethinking}            & S & 93.14 & 93.05 &  0.09 & $7.30 \times 10^{5}$ & 13.7  & $9.09 \times 10^{7}$ & 27.6 \\ 
			
			\hline
			
			ResNet-56(8-16-16)    & S & 93.15 & \textbf{92.28} & \textbf{0.67}  & $\textbf{2.70} \times \textbf{10}^{\textbf{5}}$ & \textbf{68.3}  & $5.39 \times 10^{7}$ & 57.6 \\

			\cite{DBLP:conf/iccv/HeZS17}        & F & 92.80 & 91.80 &  1.00 & $-$         & $-$     & $-$       & 50.0 \\ 
			
			\cite{ISI:000530668200024}          & F & 93.40 & 92.70 &  0.70 & $-$ & $-$  & $-$ & 50 \\
			
			\cite{DBLP:conf/cvpr/LinJYZCYHD19}     & F & 93.26 & 91.58 &  1.68 & $2.90 \times 10^{5}$ & 65.9  & $5.00 \times 10^{7}$ & 60.2 \\

			\hline 
	\end{tabular}}
	
	\caption{Comparison of pruning VGG-16 and ResNet-56 on CIFAR-10. In $``$model$"$ column, the $``$4$"$ indicates the sign of $\lambda$ is $``$—$"$, otherwise is $``$+$"$, and the value of $\beta$ is $``$4$"$. In $``$S/F$"$ column, $``$F$"$ and $``$S$"$ indicate to retrain the pruned model by finetuning or training from scratch, respectively. The $``$Acc. Drop$"$ is the accuracy of the baseline model minus that of the pruned model, so negative number means the compressed model has a higher accuracy than the baseline model. A smaller number of $``$Acc. Drop$"$ is better.}
	\label{tab1}
	
\end{table*}

\subsection{VGG-16 and ResNet-56 on CIFAR-10}

We experiment with VGG-16 and ResNet-56 on CIFAR-10. In VGG-16 pruning, we set $\beta$ to 4 and 8, respectively, and the sign of $\lambda$ is $``$-$"$. For ResNet-56 with basic blocks, we perform consistent pruning on each stage according to the characteristics of the output of the MSEB module in each layer. As shown in Figure \ref{img5}, the MSEB outputs of ResNet-56 in the first stage are relatively stable. The MSEB output of the first block in the second stage fluctuates greatly, while that of the next 8 blocks changes slightly, with 4 blocks constantly equal to 0.5. Finally, the outputs of 9 blocks in the last stage vary greatly. For stages with large fluctuations, we use a larger pruning range, and for stages with fewer fluctuations, we carefully cut or not. In the experiment, we prune the structure with three stages into 8-32-32 and 8-16-16. The specific experimental results are shown in Table \ref{tab1}.

The experiment shows that for VGG-16, our pruning method can prune the network parameters and FLOPs to a similar extent. When $\beta=8$, the compression ratio of the parameters and FLOPs in the compact VGG-16 exceed 68\%, and after retraining from scratch, the accuracy improves by 0.20\%. For MSEB-VGG-16, when $\beta=4$, the accuracy of the compact network after retraining can improve by 0.16\%, but it decreases with $\beta$ set to 8. This situation also occurs in our later experiments. We assume this may be due to overfitting by adding the MSEB module when the reduction in parameters and FLOPs in the original network influences the generalization ability to a certain extent. For ResNet-56, when the parameters and FLOPs are pruned nearly 40\% and 33.3\%, respectively, the performance after retraining improves by 0.13\%.

\begin{figure}[htbp]
	
	\centering
	
	\includegraphics[height=4.9cm,width=14cm]{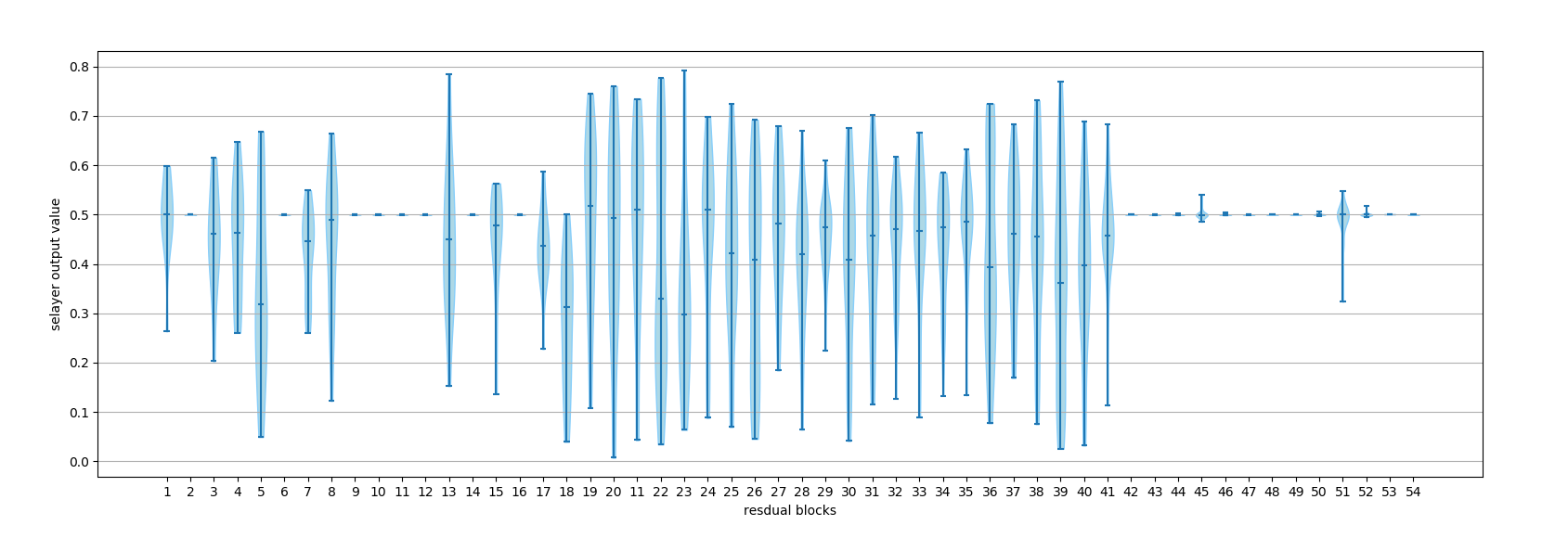} 
	
	\caption{Statistical characteristics of the MSEB output in each residual block of the PreResNet-164 network on the CIFAR-100.}
	\label{img6}
	
\end{figure}

\subsection{VGG-19 and PreResNet-164 on CIFAR-100}

\begin{table*}[htbp]
	
	\centering
	\renewcommand{\arraystretch}{1.3}
	\resizebox{\textwidth}{!}{
		\begin{tabular}{|c|c|c|c|}
			
			\hline 
			Layers  & Original Channels & Channels Pruned (3) & Channels Pruned (8)  \\ 
			\hline 
			Conv    & 64 & 40 & 35  \\ 
			Conv    & 64 & 64 & 64  \\ 
			Max Pooling    & $-$ & $-$ & $-$  \\ 
			Conv    & 128 & 128 & 128  \\ 
			Conv    & 128 & 128 & 128  \\ 
			Max Pooling    & $-$ & $-$ & $-$  \\ 			
			Conv    & 256 & 256 & 128  \\ 
			Conv    & 256 & 256 & 128  \\
			Conv    & 256 & 256 & 128  \\ 
			Conv    & 256 & 256 & 128  \\ 
			Max Pooling    & $-$ & $-$ & $-$  \\ 			
			Conv    & 512 & 256 & 256  \\
			Conv    & 512 & 133 & 129  \\ 
			Conv    & 512 & 195 & 235  \\ 
			Conv    & 512 & 256 & 394  \\ 
			Max Pooling    & $-$ & $-$ & $-$  \\ 
			Conv    & 512 & 256 & 256  \\
			Conv    & 512 & 256 & 6  \\ 
			Conv    & 512 & 256 & 230  \\
			Conv    & 512 & 256 & 104  \\ 						
			\hline 
	\end{tabular}}
	
	\caption{Comparison of the number of channels in VGG-19 before and after pruning in CIFAR-100.}
	\label{tab2}
	
\end{table*}

On CIFAR-100, we experiment with two networks, VGG-19 and PreResNet-164. For VGG-19, we set $\beta$ to 3 and 8. The number of channels in VGG-19 before and after compression is shown in Table \ref{tab2}, where the last two columns indicate the number of channels after pruning. The “3” and “8” in parentheses correspond to $\beta$. The performance of PreResNet proposed by He Kaiming et al. in \cite{DBLP:conf/eccv/HeZRS16} is exposed better than ResNet. We use our method to prune the PreResNet-164, which uses the bottleneck with $\beta$ setting to 2 and 8.

For the output characteristics of the MSEB module in each residual block of PreResNet-164, we have done data analysis, and the results are shown in Figure \ref{img6}. In the first stage, the MSEB output of some residual blocks varies greatly. In the second stage, the MSEB output of all residual blocks covers a very large range between 0-1, and the distribution is relatively on average and has no concentration. The third stage is more stable. Only a few residual blocks have slight fluctuation in MSEB output. This reveals that when performing the image classification task of the CIFAR-100, the parameters in residual blocks of the second stage of PreResNet-164 have greater redundancy than the other two stages.

\begin{table*}[htbp]
	
	\centering
	\renewcommand{\arraystretch}{2}
	\resizebox{\textwidth}{!}{
		\begin{tabular}{|c|c|c|c|c|c|c|c|c|}
			
			\hline 
			Model             & S/F & Baseline & Acc(\%)  & Acc.Drop(\%) & Parameters & Pruned(\%) & FLOPs & Pruned(\%) \\ 
			\hline 
			VGG-19            & $-$ & 72.02 & $-$     & $-$     & $2.03 \times 10^{7}$ & $-$     & $7.99 \times 10^{8}$ & $-$ \\ 
			
			MSEBVGG-19        & $-$ & 72.04 & $-$     & $-$     & $2.06 \times 10^{7}$ & $-$     & $8.00 \times 10^{8}$ & $-$ \\ 
			
			VGG-19-prund(3)     & S & 72.02 & \textbf{72.91} & \textbf{-0.73}  & $6.44 \times 10^{6}$ & 68.3  & $\textbf{4.98} \times \textbf{10}^{\textbf{8}}$ & \textbf{37.7} \\ 
			
			\cite{ISI:000530668200033}  & F & 73.26 & 73.98 & -0.72 & $-$ & 72.3  & $-$ & 34.2 \\ 
			
			\hline
			
			VGG-19-prund(8)     & S & 72.02 & \textbf{72.56} & \textbf{-0.54}  & $\textbf{3.79} \times \textbf{10}^{\textbf{6}}$ & \textbf{81.3}  & $\textbf{2.97} \times \textbf{10}^{\textbf{8}}$ & \textbf{62.8} \\

			\cite{DBLP:conf/iccv/LiuLSHYZ17}     & F & 73.26 & 73.48 & -0.22 & $5.00 \times 10^{6}$ & 75.1  & $5.01 \times 10^{8}$ & 37.1 \\ 
			
			\cite{liu2018rethinking}         & S & 72.63 & 73.08 & -0.45 & $5.00 \times 10^{6}$ & 75.1  & $5.01 \times 10^{8}$ & 37.1 \\

			\hline
			\hline 
			PreResNet-164             & $-$ & 76.90 & $-$     & $-$     & $1.73 \times 10^{6}$ & $-$     & $5.14 \times 10^{8}$ & $-$ \\ 
			
			MSEBPreResNet-164           & $-$ & 78.31 & $-$     & $-$     & $1.74 \times 10^{6}$ & $-$     & $5.15 \times 10^{8}$ & $-$ \\ 
			
			PreResNet-164(2)    & S & 76.90 & \textbf{77.02} & \textbf{-0.12} & $\textbf{1.27} \times \textbf{10}^{\textbf{6}}$ & \textbf{26.6}  & $\textbf{3.38} \times \textbf{10}^{\textbf{8}}$ & \textbf{34.2} \\ 
			\cite{liu2018rethinking}        & S & 76.99 & 77.03 & -0.04 & $1.46 \times 10^{6}$ & 13.7  & $3.44 \times 10^{8}$ & 31.1 \\ 
			\hline

			PreResNet-164(8)    & S & 76.90 & \textbf{76.47} &  \textbf{0.43} & $\textbf{7.35} \times \textbf{10}^{\textbf{5}}$ & \textbf{57.5}  & $\textbf{2.26} \times \textbf{10}^{\textbf{8}}$ & \textbf{56.0} \\

			\cite{DBLP:conf/iccv/LiuLSHYZ17}    & F & 76.63 & 76.09 &  0.54 & $1.12 \times 10^{6}$ & 34.3  & $2.25 \times 10^{8}$  & 54.9 \\ 
			
			\cite{liu2018rethinking}        & S & 76.99 & 76.02 &  0.97 & $1.12 \times 10^{6}$ & 34.3  & $2.25 \times 10^{8}$  & 54.9 \\  
			
			\hline 
	\end{tabular} }
	
	\caption{Comparison of pruning VGG-19 and PreResNet-164 on CIFAR-100.}
	\label{tab3}
	
\end{table*}

The results of VGG-19 and PreResNet-164 are shown in Table \ref{tab3}. We prune the original VGG-19, and after reducing the parameters by 68.3\% and FLOPs by 37.7\%, the accuracy of compact VGG-19 after retraining improves by 0.73. When $\beta=8$, the parameter cropping rate reaches 81.3\%, and FLOPs are reduced by 62.8\%; however, the accuracy of the compressed network is improved by approximately 0.54 compared with the original network. For PreResNet-164, our method can cut the parameters and FLOPs by approximately 26.6\% and 34.2\%, respectively, when $\beta$ is set to 2, and the final accuracy increases by 0.12.

Our UCP is to uniformly prune the channels of each layer in the CNN, including its related filters. According to different thresholds, each layer will automatically select unimportant channels to delete based on the specific situation of the feature map generated during the training process, so that the cropping rate of each layer is different. For some layers that are more important for image classification task, the cropping amplitude is relatively small, and for those layers with more redundancy, the cropping ratio will become very large. For example, the cropping ratio of the channels in the penultimate layer of VGG-19 in CIFAR-100 even exceeds 98\%. Therefore, our method can prune channels as much as possible in the image classification task, while keeping the experimental performance of the network without significant loss.

\begin{table*}[t]
	
	\centering
	\renewcommand{\arraystretch}{2}
	\resizebox{\textwidth}{!}{
		\begin{tabular}{|c|c|c|c|c|c|c|c|c|}
			
			\hline 
			Model     & Top-1 Acc(\%) & Top-1 Acc.drop(\%) & Top-5 Acc(\%)  & Top-5 Acc.drop(\%) & Parameters & Pruned(\%) & FLOPs & Pruned(\%) \\ 
			\hline 
			Resnet-18 Base   & 69.96 & $-$ & 89.42     & $-$     & 11.69M & $-$     & 1822.18M & $-$ \\ 
			
			MSEBResnet-18  & 71.10 & $-$ & 89.85 & $-$     & 11.78M & $-$     & 1823.02M & $-$ \\ 
			
			Resnet-18-prund(1)     & 69.34 & 0.62 & 88.81 & 0.61  & 8.26M & 29.34  & 1425.40M & 21.78 \\ 
			
			\cite{DBLP:conf/ijcai/HeKDFY18}  & 67.10 & 3.18 & 87.78 & 1.85  & $-$ & $-$  & $-$ & 41.80 \\ 
			
			\cite{lin2020channel}     & 67.80 & 1.86 & 88.00 & 1.08  & 9.50M & 18.72  & 968.13M & 46.94 \\
			
			\cite{lin2020channel}     & 67.28 & 2.38 & 87.67 & 1.41  & 6.60M & 43.55  & 1005.71M & 44.88 \\
			
			Resnet-18-prund(+1)    & 67.31 & 2.65 & 87.80 & 1.62  & 5.31M & 54.58  & 907.31M & 50.21 \\

			\hline
			\hline 
			Resnet-34 Base      & 73.30 & $-$ & 91.45 & $-$  & 21.80M & $-$  & 3675.63M & $-$ \\ 
			
			MSEBResnet-34       & 74.07 & $-$ & 91.68 & $-$  & 21.95M & $-$  & 3677.17M & $-$ \\ 
			
			\cite{2016arXiv160808710L}  & 72.48 & 0.75 & $-$ & $-$ & 20.10M & 7.20  & 3370M & 7.50 \\			
			
			\cite{2016arXiv160808710L}  & 72.56 & 0.67 & $-$ & $-$ & 19.90M & 7.60  & 3080M & 15.50 \\ 
			
			\cite{2016arXiv160808710L}  & 72.17 & 1.06 & $-$ & $-$ & 19.30M & 10.80  & 2760M & 24.20 \\ 
			
			\cite{8579056}  & $-$ & 0.28 & $-$ & $-$ & $-$ & 27.14  & $-$ & 27.32 \\ 		
			
			\cite{liu2018rethinking} & 73.03 & 0.28 & $-$ &  $-$ & 19.90M & 7.60  & 3080M & 15.50 \\ 
			
			\cite{liu2018rethinking} & 72.91 & 0.40 & $-$ &  $-$ & 19.30M & 10.80  & 2760M & 24.20 \\  
			
			\cite{molchanov2019taylor} & 72.83 & 0.48 & $-$ &  $-$ & $-$ & $-$  & $-$ & 24.20 \\
			
			Resnet-34-prund(2)  & \textbf{73.08} & \textbf{0.22} & \textbf{91.32} & \textbf{0.13} & \textbf{14.40M} & \textbf{33.94}  & \textbf{2562.98M} & \textbf{30.27} \\ 
			
			\cite{DBLP:conf/ijcai/HeKDFY18}  & 71.84 & 2.09 & 89.70 & 1.92 & $-$ & $-$ & $-$ & 41.10 \\				
			
			\cite{lin2020channel} & 70.45 & 2.83 & 89.69 & 1.76 & 10.52M & 51.76  & 1506.76M & 58.97 \\		
			
			Resnet-34-prund(+1) & 70.52 & 2.78 & \textbf{89.80} & \textbf{1.65} & \textbf{8.21M} & \textbf{62.34}  & 1571.40M & 57.25 \\ 	
			
			\hline 
	\end{tabular} }
	
	\caption{Comparison of pruning Resnet-18 and ResNet-34 on ILSVRC-2012.}
	\label{tab4}
	
\end{table*}

\subsection{ResNet-18 and ResNet-34 on ILSVRC-2012}

To evaluate the effectiveness of our proposed UCP method on large datasets, we experimented with ResNet-18 and ResNet-34 on the ILSVRC-2012 data set. For ResNet-18, we set the values of $\beta$ to 1 and +1, and for ResNet-34 we set the values of $\beta$ to 2 and +1. As shown in the table \ref{tab4}, the results are compared with existing methods whose results are from the original paper.

As can be seen from the table, our method is equally effective on large datasets. For ResNet-18, the baseline Top-1 classification accuracy rate is 69.69, and the Top-5 is 89.42. When the value of $\beta$ is 1, the amount of network parameters is reduced by 29.34\%, and the number of FLOPs is reduced by 21.78\%. At this time, the top-1 accuracy of the compact network on the ILSVRC-2012 drops by 0.62, and the Top-5 accuracy drops by 0.61. When the value of $\beta$ is +1, the pruning rate of parameters and FLOPs exceeds 50\%, and the Top-1 accuracy of the compressed network drops by 2.65, while the Top-5 accuracy drops by 1.62. For ResNet-34, the baseline Top-1 and Top-5 accuracy are 73.30 and 91.45 respectively. When the value of $\beta$ is 2, the amount of parameters is reduced by 33.94\%, and the FLOPs are reduced by 30.27\%. At this time, the Top-1 and Top-5 accuracy of the compact network drops by 0.22 and 0.13, respectively. When the value of $\beta$ is +1, the pruning rates of the parameters and FLOPs are 62.34\% and 57.25\%, respectively. The Top-1 and Top-5 accuracy of the compact network drops 2.78 and 1.65 respectively. Furthermore, the compact ResNet-34 has smaller parameters and FLOPs than ResNet-18, but its performance is better.

\subsection{Ablation Study}

Then, we conduct ablation analysis on the proposed UCP method. This section consists of two parts: hyperparameter study and comparison of compression ratio and accuracy drop.

\subsubsection{Hyperparameter Study}

The hyperparameters in our method include $\beta$ and the sign before $\lambda$, where the sign of $\lambda$ is used to control the relationship between the threshold and the mean value of the output sequence of the MSEB module. When it is negative, the threshold we set is less than the average value of the output of MSEB, and the pruining amplitude at this time is small. When the sign of $\lambda$ is positive, the threshold is greater than the average value, and the pruning amplitude increases. The hyperparameter $\beta$ is used to specifically control the size of the threshold. In general, the sign before $\lambda$ is equivalent to coarse control of the pruning ratio, while the value of $\beta$ is equivalent to fine control, so as to achieve different levels of pruning.
We use the experiments of PreResNet-164 on the CIFAR-100 to analyze the effects of different hyperparameters. Table \ref{tab5} shows the influence of different sign of $\lambda$ and different $\beta$ on the network compression rate, acceleration rate and performance of final compact network. The experimental setup is consistent with the previous section. The pruning rate of each layer gradually increases with $\beta$ changing. The compression rate increases from 26.6\% to 72.0\%, and the acceleration rate increases from 34.2\% to 66.1\%. As the number of channels decreases, the degree of network compression continues to increase, and the performance of the compact network on image classification datasets also gradually declines. But the loss of accuracy has been maintained within an acceptable range.

\begin{table*}[htbp]
	
	\centering
	\renewcommand{\arraystretch}{1.5}
	\resizebox{\textwidth}{!}{
		\begin{tabular}{|c|c|c|c|c|c|c|c|c|}
			
			\hline 
			Sign($\lambda$)  & Original  & \multicolumn{4}{|c|}{$-$1} & \multicolumn{3}{|c|}{$+$1}  \\ 
			\hline 
			$\beta$    & $-$ & 2 & 4 & 6 & 8 & 6 & 4 & 2   \\ 
			\hline
			Compression rate(\%)    & $-$ & 26.6 & 45.0 & 54.5 & 57.5 & 61.9 & 67.1 & 72.0  \\ 
			\hline 
			Accelerating rate(\%)    & $-$ & 34.2 & 49.6 & 54.7 & 56.0 & 58.0 & 61.3 & 66.1  \\ 
			\hline 
			Accuracy(\%)    & 76.90 & 77.02 & 76.69 & 76.53 & 76.47 & 76.21 & 75.36 & 75.49  \\ 				
			\hline 
	\end{tabular}}
	
	\caption{Pruning accuracy (\%) on CIFAR-100 dataset using different choice of hyperparameters.}
	\label{tab5}
	
\end{table*}

\subsection{Comparison of Compression Ratio and Accuracy Drop}

\begin{figure}[htbp]
	
	\centering
	
	\noindent\makebox[\textwidth][c] {
		\includegraphics[width=0.8\paperwidth]{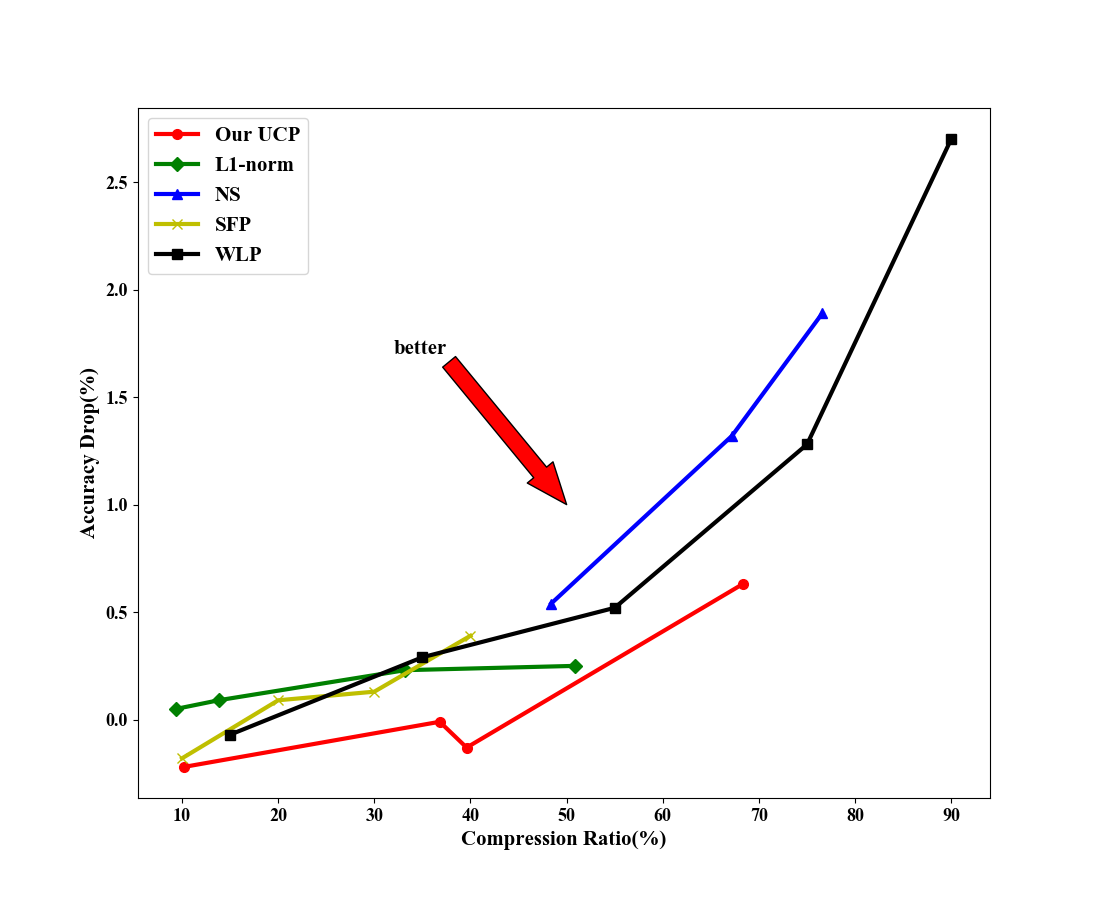} }

	\caption{Comparing the effect of compression rate on experimental accuracy drop}
	\label{img7}
	
\end{figure}

In this part, we compare our UCP method with the convolutional neural network pruning algorithms proposed in \cite{ISI:000450913101044,DBLP:conf/ijcai/HeKDFY18,DBLP:conf/iccv/LiuLSHYZ17,DBLP:conf/iccv/HeZS17}. Our method is the same as these four methods, which are structured pruning methods and based on the importance of convolution kernels or channels in the convolutional layer. Among them, Weight-Level-Pruning(WLP)\citep{ISI:000450913101044} prunes unimportant connections with small weights according to a predefined pruning rate. Soft Flter Pruning(SFP)\citep{DBLP:conf/ijcai/HeKDFY18} uses L2-norm to evaluate the importance of the filters and then prune the unimportant filters and the corresponding channels. Network Slimming(NS)\citep{DBLP:conf/iccv/LiuLSHYZ17} performs sparse induction regularization on the scaling factor of the batch normalization (BN) layer, and then prunes the channels with smaller scaling factors according to the pruning ratio. L1-norm\citep{DBLP:conf/iccv/HeZS17} uses the channel selection method based on LASSO regression to delete unimportant channels.

We compare the above methods in CIFAR-10 with ResNet-56 under the same hardware environment and parameter settings. The classification accuracy drop of each method at different compression rates is compared. The experimental results are shown in Figure \ref{img7}. It can be seen that as the network compression rate increases, the effect of our proposed method on experimental accuracy is always smaller than that of other methods. This shows that our structured uniform pruning method is more accurate and effective in evaluating the importance of channels.

\section{Conclusion and Future Work}

Deep convolutional neural networks have the best performance in many artificial intelligence applications. However, as the depth and width of CNNs increase, the large number of parameters and FLOPs limit deployment on mobile and portable devices. We propose a uniform channel pruning method that uses the MSEB module as an evaluation index of the importance for the channels to compress and accelerate CNNs. We prune the unimportant channels in the network together with the related filters. In the experiment, we prune traditional CNNs and two types of ResNet consisting of basic blocks and bottleneck. The ResNet with basic blocks is pruned consistently at the same stage based on the statistical characteristics of the output in the MSEB module to ensure the dimensional correctness of the structure. For the ResNet with bottleneck, we only prune the middle 3 $\times$ 3 convolutional layer according to the output of MSEB modules and our threshold, without changing the original dimension of every stage. Then, we retrain the compact networks from scratch and compare their performance on CIFAR-10, CIFAR-100 and ILSVRC-2012 with existing methods. The experiments show that the existing CNNs have a large redundancy of parameters and FLOPs. Our method can efficiently compress and accelerate the CNNs, and the performance of the pruned networks after retraining from scratch is better than the existing network pruning methods on the image classification dataset.

In the future, we can use the method proposed in this paper to prune GoogLeNet, DenseNet and other kind of CNNs. For ResNet, we can also combine our method with some advanced approaches for pruning intermediate channels of residual blocks to improve the compression rate.

\section*{Acknowledgments}

This work was supported in part by the National Key Research and Development Program under Grant 2018YFC0604404 and Grant 2016YFC0801804, in part by the National Natural Science Foundation of China under Grant 61806067, and in part by the Fundamental Research Funds for the Central Universities under Grant PA2019GDPK0079.

\bibliography{mybibfile}

\end{document}